\documentclass[letterpaper, 10 pt, conference]{ieeeconf}  

\IEEEoverridecommandlockouts                              

\usepackage{amsmath} 
\usepackage{graphicx}
\usepackage{caption} 
\usepackage{subcaption}
\usepackage{cite}
\usepackage[hyphens]{url}
\usepackage{comment}

\title{\LARGE \bf
 Assessing Pattern Recognition Performance of Neuronal Cultures through Accurate Simulation*
}

\author{Gabriele Lagani$^{1}$, Raffaele Mazziotti$^{2}$, Fabrizio Falchi$^{2}$, Claudio Gennaro$^{2}$, \\ Guido Marco Cicchini$^{2}$, Tommaso Pizzorusso$^{3}$, Federico Cremisi$^{4}$, Giuseppe Amato$^{2}$.
\thanks{*This work has been submitted to the IEEE for possible publication. Copyright may be transferred without notice, after which this version may no longer be accessible.}
\thanks{$^{1}$First author. Dept. of Computer Science, University of Pisa, Italy \hfill \newline
    {\tt\footnotesize gabriele.lagani@phd.unipi.it}}%
\thanks{$^{2}$CNR Pisa, Italy 
    {\tt\footnotesize raffaele.mazziotti, fabrizio.falchi, \hfill \newline claudio.gennaro, cicchini, giuseppe.amato @cnr.it}}%
\thanks{$^{3}$Dept. of Neuroscience, University of Florence, Italy \hfill \newline
    {\tt\footnotesize tommaso.pizzorusso@unifi.it}}%
\thanks{$^{4}$Scuola Normale Superiore, Pisa, Italy \hfill \newline
    {\tt\footnotesize federico.cremisi@sns.it}}%
}

\begin{document}

\maketitle
\thispagestyle{empty}
\pagestyle{empty}

\begin{abstract}
Previous work has shown that it is possible to train neuronal cultures on Multi-Electrode Arrays (MEAs), to recognize very simple patterns. However, this work was mainly focused to demonstrate that it is possible to induce plasticity in cultures, rather than performing a rigorous assessment of their pattern recognition performance. 
In this paper, we address this gap by developing a methodology that allows us to assess the performance of neuronal cultures on a learning task.
Specifically, we propose a digital model of the real cultured neuronal networks; we identify biologically plausible simulation parameters that allow us to reliably reproduce the behavior of real cultures; we use the simulated culture to perform handwritten digit recognition and rigorously evaluate its performance; we also show that it is possible to find improved simulation parameters for the specific task, which can guide the creation of real cultures.
\end{abstract}

\section{Introduction}

Interesting insights on neural network behavior can be obtained by analyzing the activity patterns of neuronal cultures 
\cite{ruaro, ferrandez, shahaf, goel}.
Multi-Electrode Arrays (MEAs) 
\cite{gross, pine} 
can be used for this purpose. MEAs are  two-dimensional electrode array plates for parallel electrophysiological recording in neuronal cultures that can be used to stimulate and read activation values (spiking activity) from cultured neuronal networks.

So far, such cultures have been used for very simple pattern recognition tasks (such as horizontal or vertical bars or simple temporal signals) \cite{ruaro, ferrandez, shahaf, goel}. 
However a rigorous assessment of their performance is lacking, and it is not clear to which extent they can be used for more complex tasks.



The contributions of this work are the following: 1) we designed a simulated model of a cultured neuronal network on a MEA; 2) simulation parameters (e.g. learning rate, excitatory/inhibitory strength, etc.) were tuned in order to reproduce the behavior of biological cultures obtained in a real-world scenario \cite{ruaro}; 3) the resulting model was trained to recognize 0 and 1 digits from the MNIST dataset \cite{mnist}, and the classification accuracy was evaluated. The model achieved 89\% accuracy in the classification task, but we were also able to improve performance up to 95\% by appropriately fine-tuning the simulation parameters. Knowledge of the resulting parameters can guide neuroscientists in the creation, through modern cultivation techniques, of real-world cultures with analogous properties \cite{eiraku, gaspard, chambers, terrigno, gonccalves, bertacchi}, optimized for the desired task.

Therefore, our methodology suggests a path of cooperation between computer science and neuroscience, in scenarios where parameter tuning in a simulated setting is more economical than in the real world, towards the future perspective of building neural engineering products.

This paper is organized as follows: 
Sect. \ref{sec:backg_relw} overviews some background and related material; 
Sect. \ref{sec:model} describes how neuronal cultures were modeled in our simulations; 
Sect. \ref{sec:simul_params} illustrates the experimental scenario that allowed us to obtain reliable simulation parameters; 
Sect. \ref{sec:digit_rec} describes the digit recognition experiments; 
the results of the various simulations are presented in Sect. \ref{sec:results} \footnote{\label{fnt:code_link} The code to reproduce the experiments is available at \newline \url{www.github.com/GabrieleLagani/SpikingGrid}}; 
finally, in Sect. \ref{sec:conclusions}, we present our conclusions and hints for future works.

\section{Background and related work} \label{sec:backg_relw}

The application of biological neural cultures for pattern recognition is a difficult challenge, because the experimenter has little or no control on the network architecture and the learning mechanism for the task at hand. In the past few years, neural cultures have been applied to the learning tasks involving only very simple patterns. 
In \cite{ruaro}, MEAs were used to stimulate a network with patterns composed of horizontal or vertical bars. The patterns were presented by sending voltages on the electrodes of the MEA in the desired positions. In particular, two patterns were constructed, one with an horizontal and a vertical bar crossing at the lower-left corner of the MEA (forming an ``L'' shape) and its transpose. The work showed that after \textit{tetanization} with the ``L'' pattern (i.e. after repeated presentation of the pattern) the network was able to respond more strongly when the same pattern was presented again, while the response to the other pattern was weaker. Thus, the network was able to distinguish between these two simple patterns. 
In \cite{shahaf}, temporally repeated stimulation was delivered to the network. It was shown that, initially, the network took about 50ms to produce a response, but, after some cycles of pattern presentation, this time became significantly shorter.
In \cite{goel}, networks were also stimulated with temporally interleaved patterns. The authors showed that specific changes occurred in the network connectivity, which allowed the network to learn temporal intervals for predicting when the next stimulus would occur.

Spiking Neural Networks (SNNs) are a realistic model of biological networks 
\cite{gerstner}. 
In SNNs, neurons communicate by short pulses called \textit{spikes}. All the spikes are equal to each others and values are encoded in the timing or frequency with which they are emitted. Neurons are modeled as Leaky Integrate and Fire (LIF) units
: they sum up all the received spikes (weighted by the synaptic coefficients) and when this sum, represented by the neuron membrane potential, exceeds a threshold, an output spike is emitted. At these point the membrane potential is reset and the neuron cannot spike anymore for a period called \textit{refractory time}. These units are leaky in the sense that, when no spikes are received in input, the membrane potential decays exponentially. Synaptic modification occurs by Spike Time Dependent Plasticity (STDP) 
\cite{gerstner}. 
In STDP, weight strengthening occurs when an input spike on a given synapse is immediately followed by an output spike. Hence, when an input spike is a potential cause for the output firing, STDP reinforces this correlation. When, instead, an input spike is immediately preceded by an output spike, the weight on the input synapse is weakened. Formally, the STDP learning rule can be expressed as:
\begin{equation} \label{eq:stdp}
    \Delta w = 
    \begin{cases}
        A_+ e^{-(t_{out} - t_{in}) / \tau_+} & \text{if $t_{out}   >  t_{in}$}\\
        A_- e^{ (t_{out} - t_{in}) / \tau_-} & \text{if $t_{out} \leq t_{in}$}
    \end{cases}
\end{equation}
where $A_+$ and $A_-$ are the learning rate parameters for potentiation and weakening respectively, $tau_+$ and $tau_-$ are time constant parameters which determine the effect of input and output spikes based on their temporal distance, $t_{in}$ is the input spike time and $t_{out}$ is the output spike time. 



\section{Modeling neuronal cultures} \label{sec:model}

We used the SNN model to build a digital counterpart of the cultures used in \cite{ruaro}. We considered a MEA composed of 60 electrodes, disposed in a grid of 6x10 elements, with dimensions 3mm x 5mm. About 10000 neurons (each receiving approximately 3000 excitatory and 500 inhibitory connections) were randomly placed on top of the grid, forming a SNN with random connectivity. Connectivity was set up randomly, with neighboring neurons having a high probability of connecting with each other, and distant neurons having a small probability of forming a connection. The connection probability decreased with distance following a Gaussian profile. The standard deviation parameter of the Gaussian profile allowed to control the connectivity range. We used different Gaussian profiles for the excitatory and inhibitory connections (parametrized by standard deviations $\sigma_E$ and $\sigma_I$). 
Connection weights were initialized to small random values. More precisely, weight values were drawn from a uniform distribution between 0 and 1, and then re-scaled by a factor controlled by simulation parameters (EXC\_STRENGTH and INH\_STRENGTH).
Stimuli were delivered to neurons through voltages produced by the MEA electrodes. Plasticity of the connection weights occurred according to the STDP eq. \ref{eq:stdp}, for excitatory connections, and to anti-STDP (i.e. the same rule but with opposite sign) for inhibitory connections.
The complete list of simulation parameters is shown in Tab. \ref{tab:bio_params}.
The values for these parameters were determined as discussed in the next section.

\section{Determination of simulation parameters} \label{sec:simul_params}

\begin{figure}[t]
\centering 
\begin{subfigure}{\linewidth}
\centering
\includegraphics[width=0.5\linewidth]{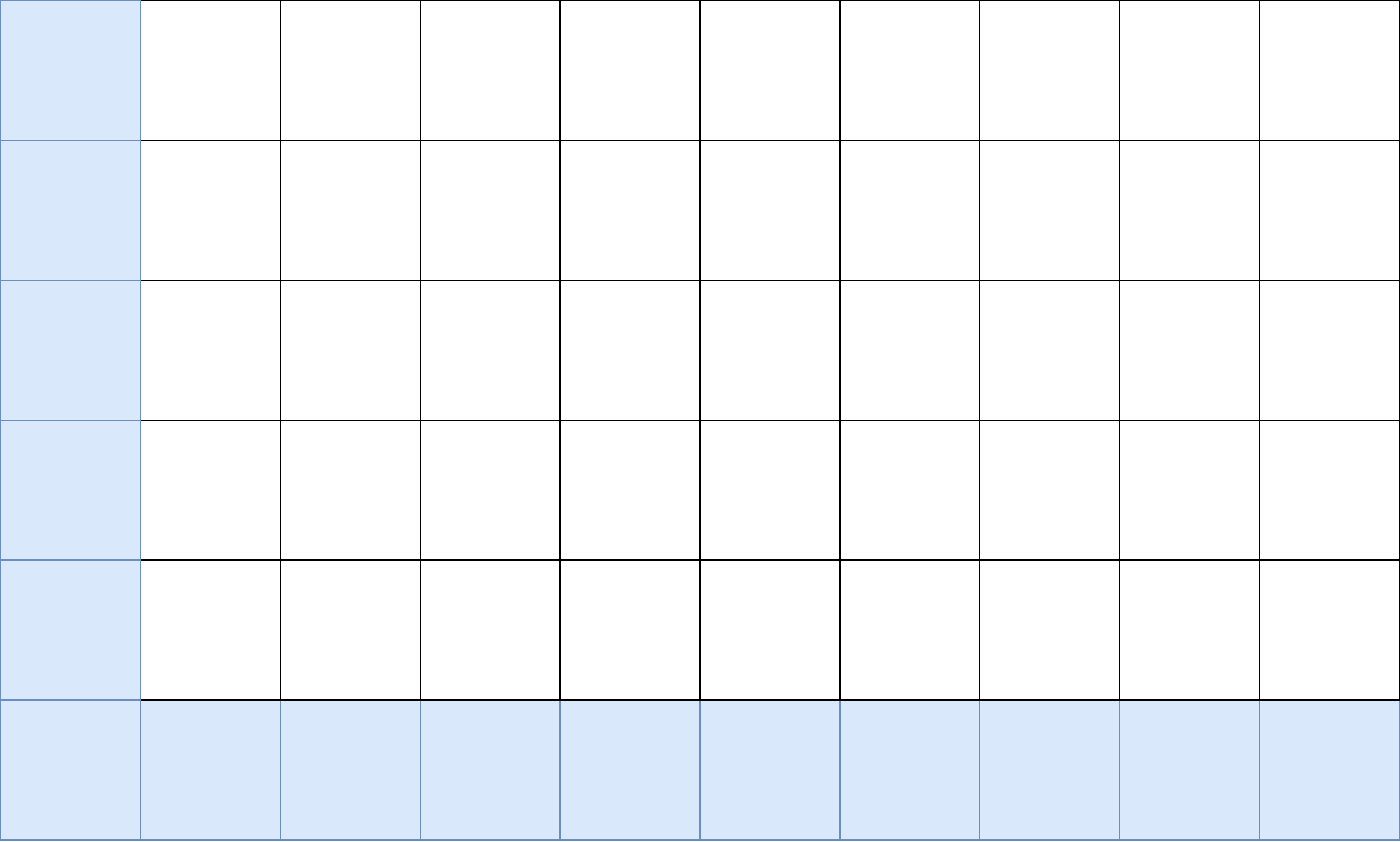}
\caption{Example of MEA electrode activations forming an L-shaped pattern.}
\label{fig:grid_L}
\end{subfigure}
\\
\begin{subfigure}{\linewidth}
\centering
\includegraphics[width=0.5\linewidth]{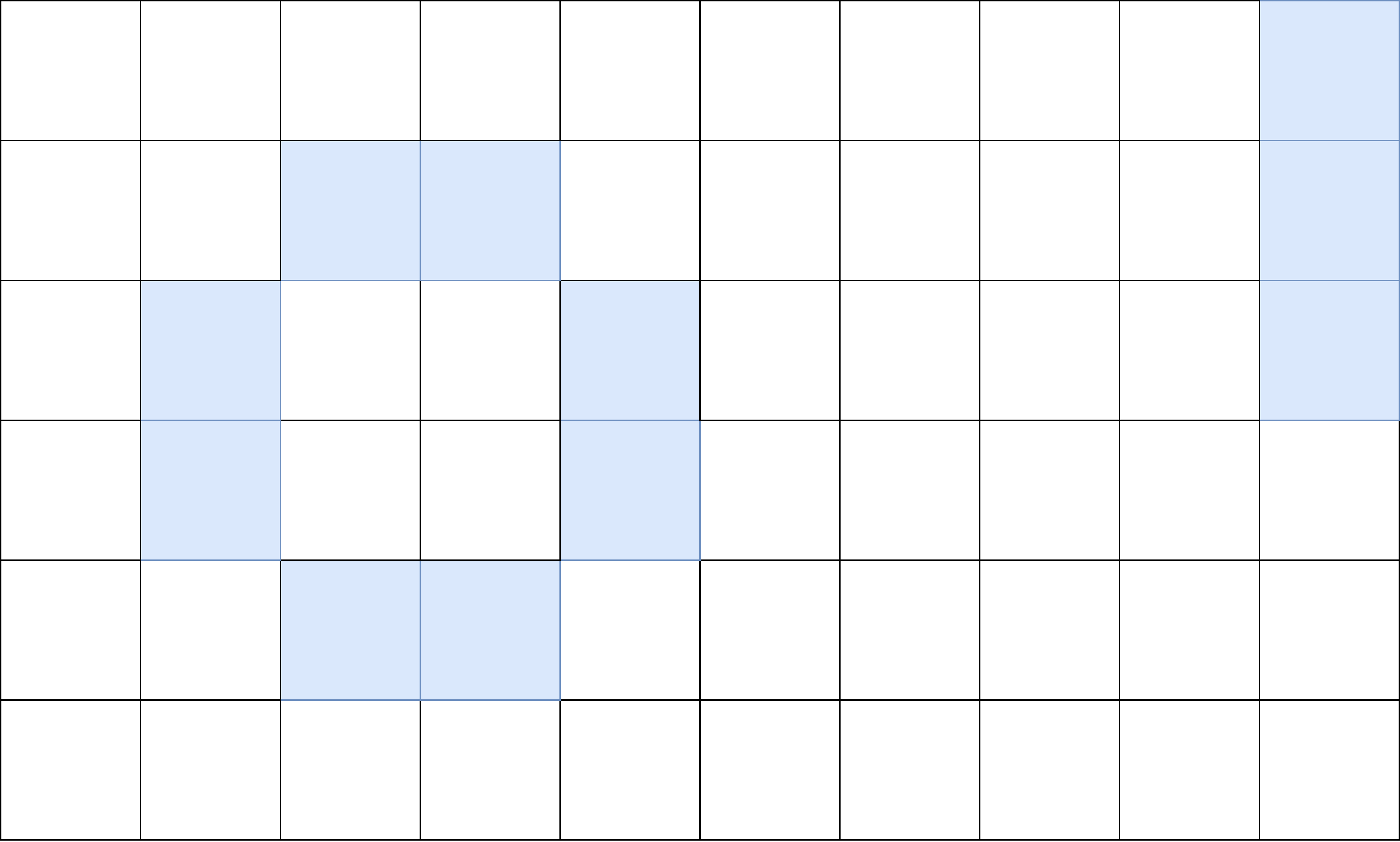}
\caption{Example of MEA electrode activations during training for the digit recognition task. The digit image is encoded in the 6x6 region of electrodes on the left, while, at the same time, the class label stimulation is delivered on the electrodes on the right.}
\label{fig:grid_digits}
\end{subfigure}
\caption{6x10 MEA grid with activated electrodes forming the desired input stimulation.}
\label{fig:grid}
\end{figure}

Our experiments evolved in two stages. In the first stage, we reproduced the experimental scenario described in \cite{ruaro}, and we tuned the simulation parameters in order to obtain a behavior of the simulated culture as close as possible to that of real-world cultures. In this scenario, the network was stimulated with two types of patterns: an L-shaped pattern, as shown in Fig. \ref{fig:grid_L}, and its transpose. Stimulation consisted in generating pulses on the desired electrodes of the MEA. By virtue of the STDP updates, repeated stimulation with the same pattern caused the network to consolidate its weights in order to produce a stronger response to that pattern in the future. This process is known as \textit{tetanization}. We performed tetanization using the L-shaped pattern, which was repeatedly presented to the network as a spike train of 100 pulses at a rate of 250Hz, thus having a total duration of the spike train of 400ms. A total of 40 spike trains were delivered to the network during the whole tetanization process. The response of the network to the L-shaped pattern and its transpose was evaluated both before and after tetanization.

We tuned simulation parameters by performing a grid search within biologically plausible ranges \cite{jug}. Optimal parameters were those minimizing the Mean Squared Error (MSE) between the temporal evolution of the response (averaged over all neurons) of the artificial and real cultures following a stimulation. 

\section{Digit recognition experiment} \label{sec:digit_rec}

In the second stage of our experiments, we used the neuronal culture for recognizing 0 and 1 digits from the MNIST dataset \cite{mnist}. 
Digit images were resized to 6x6 pixels, in order to fit them to the MEA grid;  then, each image was mapped to the 6x6 sub-grid of electrodes on the left part of the 6x10 MEA. 
Stimulation on each electrode consisted of trains of pulses at a rate proportional to the intensity of the corresponding pixel in the image, up to 200Hz, for 100ms.
In order to allow the network to learn input-class mappings, we associated some of the neurons in the remaining 6x4 part of the MEA, where no image pixels were mapped, with class labels. We call the neurons associated with class labels \textit{output neurons}. When an input was presented to the network, the number of spikes emitted by neurons associated with each class were counted, and the predicted class was chosen as that associated with the group of neurons that spiked the most. In order to enable learning of input-class mappings, a variant of the \textit{teacher neuron} technique \cite{shrestha} was designed: during training, when an input of a given class was presented, the electrodes on the locations of the grid associated with the given class were also activated in order to deliver a high frequency stimulation, at 200Hz, to output neurons, simultaneous with input presentation. This caused the output neurons to fire strongly, thus developing input-class associations through the STDP rule. More in detail, neurons placed on the electrodes in the last column and first three rows were associated with class 0, while neurons placed on electrodes in the last column and last three rows were associated with class 1. Fig. \ref{fig:grid_digits} shows an illustration of electrode activation for an example digit presentation, together with the activation of the class label electrodes, for the training task on the digit recognition dataset. At test time, the label-related stimulation was removed, and the output neurons had to reconstruct the missing label depending on the input presented. The simulations were implemented in Python, using the Bindsnet package \cite{bindsnet} for SNN simulation.

\section{Results} \label{sec:results}

\begin{figure*}[t] 
\centering 
\begin{subfigure}{0.3\textwidth}
\centering
\includegraphics[width=\textwidth]{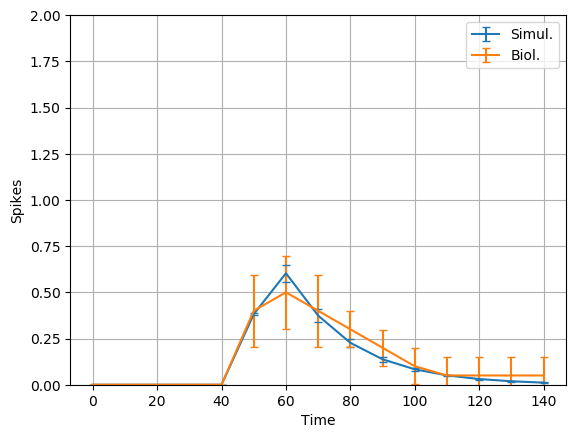}
\caption{Temporal response to regL pattern, before tetanization.}
\label{fig:time_resp_a}
\end{subfigure}
~
\begin{subfigure}{0.3\textwidth}
\centering
\includegraphics[width=\textwidth]{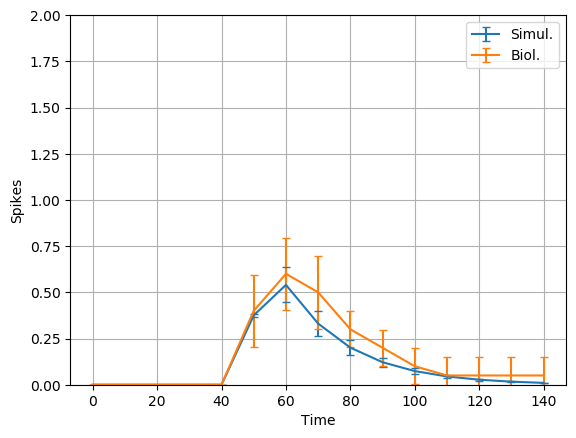}
\caption{Temporal response to upsL pattern, before tetanization.}
\label{fig:time_resp_b}
\end{subfigure}
\\
\begin{subfigure}{0.3\textwidth}
\centering
\includegraphics[width=\textwidth]{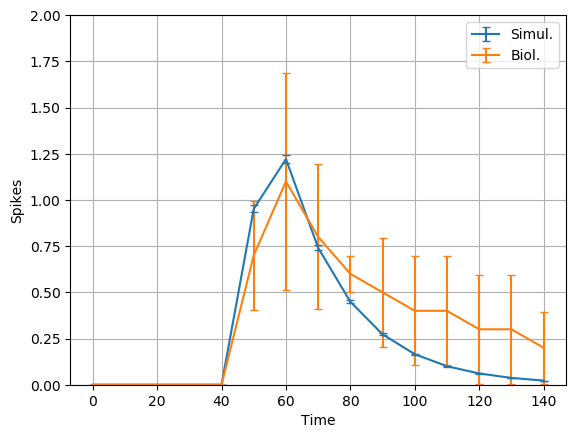}
\caption{Temporal response to regL pattern, after tetanization.}
\label{fig:time_resp_c}
\end{subfigure}
~
\begin{subfigure}{0.3\textwidth}
\centering
\includegraphics[width=\textwidth]{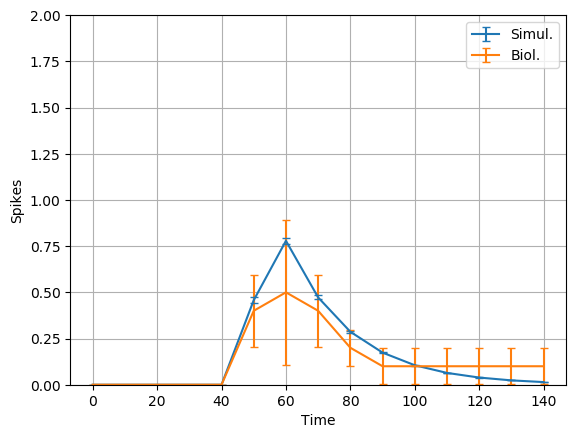}
\caption{Temporal response to upsL pattern, after tetanization.}
\label{fig:time_resp_d}
\end{subfigure}
\caption{Temporal response to regular L (regL) and upside-down L (upsL) patterns, before and after tetanization. Comparison of simulation and real-world (biological) data.}
\label{fig:time_resp}
\end{figure*}

For the first stage of experiments, we evaluated the spatio-temporal response of the network to both regular L-shaped pattern (named \textbf{regL}) and to the upside-down (i.e. the transposed) L pattern (named \textbf{upsL}). We collected data from four independent repetitions of the experiments (because also the biological experiment was repeated using four different neuronal cultures) and we averaged the results and computed 95\% confidence intervals. In Fig. \ref{fig:time_resp}, the temporal response, in terms of number of spikes emitted, obtained by averaging the spike count over all the neurons is shown. Spikes counts are computed over time bins of 10ms each, and the response over a time period of 150ms is plotted, comparing the simulated results with real-world data. Stimulus delivery occurs at 50ms, and from this time instant we observe a sudden response which then decays over time. Moreover, after tetanization, we observe that the strength of the response to regL patterns increases w.r.t. that of upsL patterns. 

The first experimental stage gave us a set of simulation parameters that allowed to obtain results most similar to real-world data (Tab. \ref{tab:bio_params}). For the second stage of experiments, we evaluated the performance achieved by the neuronal culture, after training, in the digit recognition task, obtaining 89\% accuracy. For the simulation, we used the parameters from the first experimental stage. We were also able to further improve the accuracy up to 95\%, by appropriately fine tuning some of the simulation parameters. 
We found that the parameters related to excitatory and inhibitory connection strength had a great impact on the final results. In particular, we found that we could improve the results by increasing the excitatory connection strength by a factor of 8, and further increasing the inhibitory connection strength to be two orders of magnitude larger than the excitatory strength. Such strong inhibitory connections can be justified in that they allow to realize shunting inhibition, resulting in competitive interaction among neurons, which is at the basis of \textit{competitive learning} 
\cite{grossberg}.

\begin{table}[t]
\centering
\caption{Simulation parameters that produced simulated response most similar to the real-world data}
\begin{tabular}{|c|c|}
\hline
\textbf{Parameter} & \textbf{Value} \\
\hline
Resting membrane potential & -70mV \\
\hline
Threshold membrane potential & -50mV \\
\hline
Reset membrane potential & -70mV\\
\hline
Refractory period & 5ms \\
\hline
Membrane potential decay time & 50ms\\
\hline
STDP trace decay time & $\tau_+ = \tau_- = 20ms$ \\
\hline
Learning rate & $A_+ = 10^{-2}, A_- = 10^{-4}$\\
\hline
EXC\_STRENGTH & 1 \\
\hline
INH\_STRENGTH & 10 \\
\hline
$\sigma_E$ & 1.2mm \\
\hline
$\sigma_I$ & 0.15mm \\
\hline
\end{tabular}
\label{tab:bio_params}
\end{table}

\section{Conclusions and future work} \label{sec:conclusions}

In this paper, we explored the possibility of using neuronal cultures for pattern recognition. We developed a simulator of neuronal cultures on MEA devices. We compared the behavior of the simulated culture with that of biological cultures, tuning the simulation parameters to make the simulated results as close as possible to the real-world data. Then, using the same simulation parameters, we trained a simulated neuronal culture to recognize 0 and 1 digits, obtaining an accuracy of 89\%. We also found that, by appropriately modifying the simulation parameters, it was possible to further improve the accuracy to 95\%.

We have shown that, through simulation, it is possible to obtain insights on the parameters and properties (such as strength and range of excitatory and inhibitory connections) that a neuronal culture should have in order perform well at a given task. These insights can then be used by neuroscientists in order to develop biological networks, by means of modern cultivation techniques, with the desired properties \cite{eiraku, gaspard, chambers, terrigno, gonccalves, bertacchi}. 

In the future, we plan to reproduce the experimental scenario of 0 and 1 recognition also \textit{in vitro}. We also propose to move towards more complex tasks, for instance, recognition of more digits or even natural images. For this purpose, it might be useful to consider larger MEAs (e.g. with 4096 electrodes). Also, for more complex tasks, it might be necessary to consider a different learning approach, such as reward-driven \cite{williams, florian, skorheim}. 



\bibliographystyle{IEEEtran}
\bibliography{references}

\end{document}